\documentclass[published]{agujournal2025}
\usepackage{subcaption}
\usepackage{listings}
\usepackage[utf8]{inputenc}
\usepackage[T1]{fontenc}
\usepackage{xcolor}

\begin{document}


\title{Physics-informed reservoir characterization from bulk and extreme pressure events with a differentiable simulator}

\authors{%
                  Harun Ur Rashid\affil{1},
                  Mingxin Li\affil{2}, 
                  Aleksandra Pachalieva\affil{1},
                  Georg Stadler\affil{2},
                  Daniel O'Malley\affil{1}
	      }

\affiliation{1}{Earth and Environmental Sciences Division, Los Alamos National Laboratory, Los Alamos, NM, 87545, USA}
\affiliation{2}{Courant Institute School of Mathematics, Computing and Data Science,
New York University, New York City, NY 10012, USA}

\authoraddr{%
                      Harun Ur Rashid, Earth and Environmental Sciences Division, Los Alamos National Laboratory, Los Alamos, NM, 87545, USA, (hrashid.lsu@gmail.com).
                   }

\authorrunninghead{Ignored.} 

%
\keypoints
{We developed a physics-informed ML method for reservoir characterization that consistently outperforms purely data-driven approaches.}
{The proposed method infers heterogeneous reservoirs from extreme pressure data more accurately, improving risk-aware characterization.}
{Our approach links fast surrogate inferences and physics for real time subsurface characterization in bulk and rarely observed data regimes.}


\titlerunninghead{Ignored.}

\maketitle

\bigskip  

\begin{abstract}
Accurate characterization of subsurface heterogeneity is challenging but essential for applications such as reservoir pressure management, geothermal energy extraction and CO$_2$, H$_2$, and wastewater injection operations. This challenge becomes especially acute in extreme pressure events, which are rarely observed but can strongly affect operational risk. Traditional history matching and inversion techniques rely on expensive full-physics simulations, making it infeasible to handle uncertainty and extreme events at scale. Purely data-driven models often struggle to maintain physics consistency when dealing with sparse observations, complex geology, and extreme events. To overcome these limitations, we introduce a physics-informed machine learning method that embeds a differentiable subsurface flow simulator directly into neural network training. The network infers heterogeneous permeability fields from limited pressure observations, while training minimizes both permeability and pressure losses through the simulator, enforcing physical consistency. Because the simulator is used only during training, inference remains fast once the model is learned. In an initial test, the proposed method reduces the pressure inference error by half compared with a purely data-driven approach. We then extend the test over eight distinct data scenarios, and in every case, our method produces significantly lower pressure inference errors than the purely data-driven model.  We also evaluate our method on extreme events, which represent high-consequence data in the tail of the sample distribution. Similar to the bulk distribution, the physics-informed model maintains higher pressure inference accuracy in the extreme event regimes. Overall, the proposed method enables rapid, physics-consistent subsurface inversion for real-time reservoir characterization and risk-aware decision-making.
\end{abstract}
\section{Introduction}  
Reservoir characterization is the process of quantitative estimation of subsurface reservoir properties such as porosity, permeability, and fluid saturation based on available observation data, geological models, and fluid-flow behavior \cite{lake2012reservoir}. The goal of reservoir characterization is to build a quantitative reservoir model based on the extracted information to reliably infer flow and transport behavior \cite{lucia2003carbonate} within the reservoir.

Accurate reservoir characterization is essential for optimizing injection and production operations \cite{yang2003integrated,baker2015practical} , manage reservoir pressure \cite{harp2021feasibility, pachalieva2022physics}, and reduce risks such as leakage and induced seismicity \cite{goertz2017characterization}, thus supporting informed decision-making throughout reservoir development and operation. This need is especially critical in the geological storage of CO$_2$, where reservoir heterogeneity strongly controls pressure buildup, plume migration, trapping behavior and long-term containment integrity. Therefore, poor characterization can lead to overpressure, leakage, and inefficient use of storage capacity, ultimately compromising sequestration goals \cite{chadwick2004geological,forster2010reservoir,sundal2016lower,medina2017characterization,fawad2021seismic}. However, reliable characterization remains difficult because subsurface reservoirs are highly heterogeneous, uncertain, and governed by multiscale, multiphysics processes, while information about their internal structure is typically sparse and uncertain. As characterization must frequently rely on sparse data from a few exploratory wells, modeling tools must balance speed with physical accuracy. However, achieving this balance of capturing essential physics while maintaining computational efficiency under data constraints remains challenging. To date, no such robust and fast characterization model exists.

Traditional reservoir characterization typically involves history matching, where physics-based reservoir simulators are calibrated to reproduce observed production or pressure data \cite{thomas1972nonlinear,wasserman1975practical,williams1998stratigraphie,oliver2011recent,rwechungura2011advanced,christie2013use,li2023history}. While full-physics simulators provide high fidelity representations of subsurface processes, they are computationally intensive, making inverse modeling and risk assessment time consuming and often impractical.

Another conventional approach is inverse modeling, particularly within a Bayesian framework, which incorporates prior geological knowledge and accounts for observational uncertainty to generate probabilistic estimates of reservoir parameters \cite{haldorsen1993challenges,gunning2007detection,grana2015bayesian,qin2018petrophysical,grana2022probabilistic}. These methods often rely on adjoint-based gradient computations to efficiently evaluate sensitivities of model outputs with respect to input parameters. While scalable and effective in data-sparse settings, Bayesian inversion can become prohibitively expensive for large, complex reservoirs, especially when employing sampling-based techniques such as Markov Chain Monte Carlo (MCMC), which require thousands of model evaluations for convergence \cite{gallagher2009markov}.

Recent advances in reservoir characterization have increasingly adopted purely data-driven inverse modeling approaches, where machine learning models, including neural networks and convolutional neural networks (CNNs), are trained on pre-generated datasets to learn a direct mapping from observed data to subsurface properties such as permeability fields \cite{helle2001porosity, zhou2019data,wang2021ensemble,liu2021extreme,okon2021artificial}. Once trained, these models can provide rapid estimates without the repeated forward simulations required in conventional full-physics inverse modeling. However, as they effectively learn statistical interpolation rules from the training set, their predictive capability often degrades when applied to geological scenarios outside the training distribution. Moreover, they generally do not enforce the governing flow equations, which can lead to physically inconsistent inference. Their success therefore depends heavily on access to large, diverse, and representative training datasets, which are costly to generate and often unavailable in practical applications.

To address the limitations of purely data-driven inverse models, physics-informed approaches such as physics-informed neural networks (PINNs) have emerged as an alternative framework. Rather than learning a direct mapping from observed data to reservoir parameters, PINNs typically represent the solution of the governing flow equations with a neural network and train that network by minimizing a loss that includes both data mismatch and the residuals of the underlying partial differential equations (PDEs) \cite{shokouhi2021physics,lv2021novel,behl2023data,nagao2024physics,li2024probabilistic,khassaf2025physics}. This optimization-based formulation can, in principle, be used to infer unknown reservoir properties simultaneously with the state variables, while incorporating the governing physics as a soft constraint through the PDE residual term.
However, in subsurface flow applications, PINNs and related neural PDE solvers often face significant difficulties when the system exhibits strong heterogeneity, sharp spatial gradients, or highly nonlinear multiphase behavior \cite{wang2024solving}. Under such conditions, accurately representing the forward solution with a neural network can be challenging, which may reduce robustness and accuracy compared with established numerical discretization methods.

To address the limitations of existing reservoir characterization methods, a more advanced alternative to physics-constrained surrogates involves embedding a fully differentiable, full-physics simulator directly into the machine learning training loop. This enables physics-aware learning for efficient and accurate reservoir characterization. For seamless integration into the training process, the simulator must be fully differentiable. However, conventional numerical simulators \cite{pruess1991tough2, pettersen2006basics, rashid2022iteratively} are not inherently differentiable and therefore incompatible with gradient-based optimization. To enable such integration, subsurface flow simulators can be built using differentiable programming \cite{innes2019differentiable} and automatic differentiation (AD) techniques. These tools allow gradients to be computed via backpropagation through the simulator, which is  an essential requirement for optimizing surrogate models within frameworks such as PyTorch \cite{ketkar2021introduction} and Flux \cite{innes2018flux}. 

By integrating a differentiable full-physics simulator directly into the training loop, we gain greater control over the learning process and explicitly guide the model to capture complex relationships between input and output variables. In this setup, the neural network outputs are fed into the physics simulator, which generates inferences that can be compared with observed data. The discrepancy between simulated and observed responses is then incorporated into the loss function, along with the neural network's inference errors. This physics-guided learning approach enhances inference accuracy, even in scenarios with sparse and rare observational data. Recent efforts in physics-guided machine learning include the use of model-constrained autoencoder frameworks \cite{jin2020physics, pakravan2021solving, van2025taen}. Rather than using a purely data-driven encoder–decoder, these encoders incorporate the governing model into training so that the learned latent representation and decoded output are constrained by the underlying PDE. However, such approaches have not yet been developed for reservoir characterization problems involving strongly heterogeneous formations and complex subsurface flow physics.

In this study, we introduce a data driven, physics-informed  machine learning method (which we refer to as the physics-informed method in the rest of this paper) for characterizing heterogeneous reservoirs using pressure data. We train an artificial neural network using randomly generated permeability samples, which infers formation permeability based on pressure at fixed observation points. During the training, in addition to the permeability mismatch in the purely data-driven model (which we  refer to as the data-driven method in the rest of this paper), our physics-informed model accounts for the pressure mismatch in the loss function. We calculate this pressure loss by passing the inferred permeability into a fully differentiable flow simulator. Across the scenarios considered, the proposed approach reduces pressure-inference error by 33\% to 64\% relative to the data-driven model. The key contributions of this work are: (i) We develop a physics-informed machine learning method for reservoir characterization by embedding a fully differentiable simulator within the training loop of surrogate models. (ii) We apply the proposed method to infer heterogeneous reservoir permeability from extreme and thus rarely observed data to improve risk-aware characterization. 

The remainder of this work is organized as follows: We first describe our methods in detail, then present our results, and conclude with a discussion of the findings of this research. In Section \ref{sec:method}, we review the background physics, physics-based simulation setup and NN model architecture. We also present training methods for both physics-informed and data-driven approaches. In Section \ref{sec:results}, we present the training and validation outcomes for the base case and the outcomes for different data scenarios. Here, we also present training and inference results on extreme events.

In Section \ref{sec:discussion}, we highlight the strengths and limitations of the proposed framework and outline directions for future improvement. Finally, we conclude by summarizing the novelty, strengths, applicability, and limitations of our approach.

\section{Method}\label{sec:method}
In this section, we describe the full methodology used in our study. We begin by presenting the governing equations of the full physics simulator, the problem setup, computational parameters, and solution procedure. We then introduce the architecture of the neural network model and explain the parameters and data structures used for training and validation. Finally, we outline the training methods for our data-driven and physics-informed model. 

\subsection{Physics model}
Our physics model connecting subsurface permeability with  pressure data is the steady-state, single-phase Darcy flow model, described in detail below. Solving this equation maps the permeability field to pressure point data at monitoring points. Our aim is to construct a map, parameterized by a neural network, that goes in the reverse direction, i.e., from
pressure point data to (an estimate of) the permeability field. One of the methods we propose to learn such a map requires differentiating through the (discretized) Darcy flow equations. Thus, it is convenient to solve this equation using DPFEHM, an open-source Julia package for subsurface flow modeling \cite{o2023dpfehm} that provides derivatives through Julia’s automatic differentiation capability.

The single-phase steady-state model assumes that pressure changes caused by fluid injection or extraction that occur within a system are independent of time \cite{von1956mechanics}. In a heterogeneous reservoir, this behavior is described by the following partial differential equation: 
\begin{equation}
    \nabla \cdot \left( K(x) \nabla p \right) = q,
\end{equation}
where $p$ is the pressure, $K(x)$ is the scalar spatially varying permeability field, and $q$ represents sources and sinks.

\begin{figure}[!h]
\centering
\includegraphics[width=0.60\linewidth]{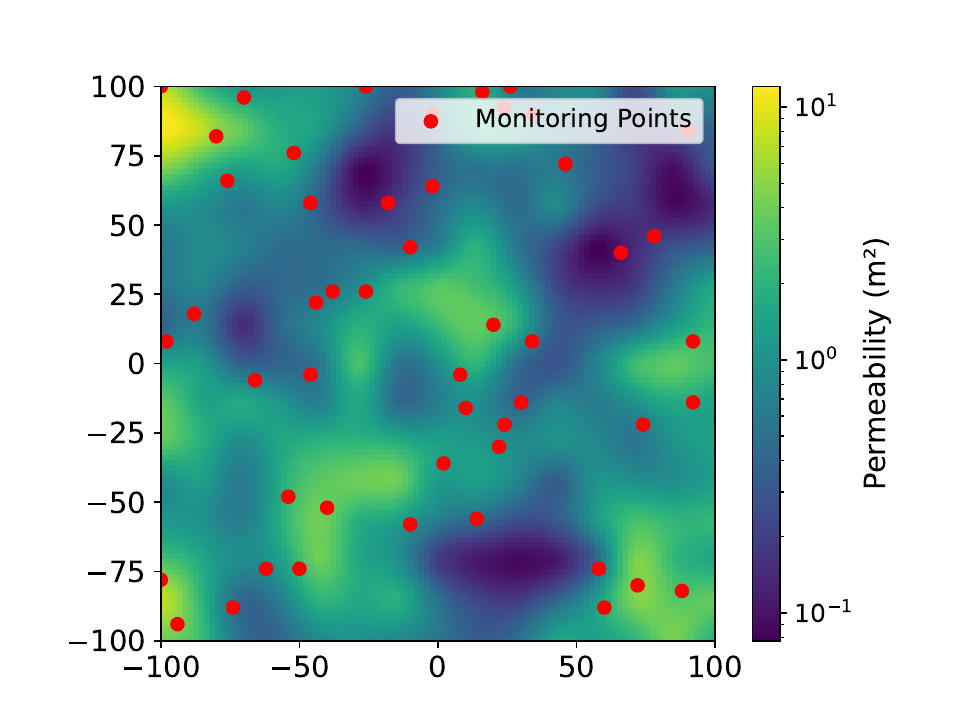}
\caption{Simulation domain used for the reservoir characterization method with monitoring points in red and a heterogeneous permeability field in the background. The trained NN model model infers the permeability of the field from the pressure values at the monitoring points.}
\label{fig:problemsetup}
\end{figure}

We consider a two-dimensional reservoir domain to define the physical problem underlying our methods. Figure \ref{fig:problemsetup} illustrates the computational domain, which is a \(200\,\mathrm{m} \times 200\,\mathrm{m}\) square. Dirichlet pressure boundary conditions are imposed on all boundaries. Pressure observations are collected at a set of randomly selected monitoring locations within the domain, shown as red dots in Figure \ref{fig:problemsetup}. To represent geological heterogeneity, the permeability is modeled as a spatially varying random field. This field is parameterized using a truncated Karhunen--Lo\`eve (KL) expansion derived from a prescribed covariance model for the log-permeability field. In particular, we assume a Matérn-type covariance structure with a specified correlation length, so that each permeability realization is represented by a finite-dimensional coefficient vector, which determines the corresponding heterogeneous permeability field. These KL coefficients serve as input to the forward model, which maps them to the associated pressure response at  monitoring locations.

Using this physical setup, we construct datasets used for model training and evaluation. For each sample \(j\), a coefficient vector \(\boldsymbol{k}^{(j)}\) is drawn to define a heterogeneous permeability field \(K^{(j)}(x)\) through the truncated KL expansion. This permeability field is then passed through the forward model to compute the corresponding pressure response at the monitoring locations, denoted by \(\tilde{\boldsymbol{p}}^{(j)}\). To represent measurement uncertainty, we perturb the pressure observations in the training and validation datasets with 10\% Gaussian white noise, yielding noisy observations \(\boldsymbol{p}^{(j)}\). The resulting training and validation datasets therefore consist of paired samples \((\boldsymbol{k}^{(j)}, \boldsymbol{p}^{(j)})\). A large subset of these samples is used for training, while the remainder is reserved for validation. An additional evaluation dataset is generated from the same physical setup; in this case, the permeability field \(K^{(j)}(x)\) is treated as the reference permeability field, and the corresponding pressure response \(\tilde{\boldsymbol{p}}^{(j)}\) is used as the reference pressure data for model assessment. The objective of the method is to train a model that can recover the underlying permeability representation \(\boldsymbol{k}^{(j)}\), and hence \(K^{(j)}(x)\), from the associated pressure observations.

\subsection{Neural network model}
With the forward model \(\mathcal{F}\) and the dataset pairs \((\boldsymbol{k}^{(j)}, \boldsymbol{p}^{(j)})\) defined, we next introduce the neural network used to approximate the inverse map from pressure observations to the permeability parameterization. The goal is to learn the mapping from the pressure response measured at the monitoring locations to the truncated Karhunen--Lo\`eve (KL) coefficients that define the heterogeneous permeability field. To this end, we define a fully connected neural network
\[
\mathcal{G}_{\Theta} : \mathbb{R}^{n_{\mathrm{obs}}} \rightarrow \mathbb{R}^{n_{\mathrm{KL}}},
\]
where \(\Theta\) denotes the trainable weights and biases, \(n_{\mathrm{obs}}\) is the number of monitoring locations, and \(n_{\mathrm{KL}}\) is the number of retained KL coefficients. For each sample \(j\), the input to the network is the pressure observation vector \(\boldsymbol{p}^{(j)} \in \mathbb{R}^{n_{\mathrm{obs}}}\), and the output is a inferred coefficient vector
\[
\widehat{\boldsymbol{k}}^{(j)} = \mathcal{G}_{\Theta}\!\left(\boldsymbol{p}^{(j)}\right) \in \mathbb{R}^{n_{\mathrm{KL}}}.
\]
The inferred coefficient vector \(\widehat{\boldsymbol{k}}^{(j)}\) defines the reconstructed permeability field through the truncated KL expansion. This formulation can also be viewed as amortized inversion. Instead of solving an inverse problem for each individual set of pressure observations, we train a single neural network to learn a reusable map from observations to the KL parameterization of permeability. The computational cost is concentrated in an offline training stage, after which inference for new observation sets is fast. This is a key distinction from conventional least-squares and Bayesian inversion approaches, which usually require solving a separate conditioning problem for each dataset \cite{cranmer2020frontier}.

In this work, \(\mathcal{G}_{\Theta}\) is implemented as multilayer perceptron (MLP) with an input dimension of \(n_{\mathrm{obs}}\). The MLP has two fully connected hidden layers, each containing 64 neurons with ReLU activation functions. The input dimension corresponds to the number of pressure observations at the monitoring locations, and the output dimension corresponds to the number of KL coefficients used to represent the permeability field. This architecture provides a compact parameterization of the inverse problem: instead of directly inferring permeability at every spatial location, the network infers a low-dimensional set of coefficients that captures the dominant spatial variability of the heterogeneous field. The relatively small input dimension makes dense layers computationally efficient, while the hidden layers provide sufficient expressive capacity to approximate the nonlinear relationship between the observed pressures and the underlying permeability structure. The neural network parameters used in this study are summarized in Table~\ref{tab:param_NN}. 

\begin{table}[!h]
\caption{Parameters used in training the neural network model.}
\centering
\begin{tabular}{l c}
    \hline
    Parameter & Value \\
    \hline
    Batch size & 5 \\
    Number of batches & 100 \\
    Samples per iteration & 500 \\
    Number of iterations & 10000 \\
    Total number of training samples & [5000, 50000] \\
    Number of validation samples & 200 \\
    Optimizer & ADAM \\
    Learning rate & \(10^{-4}\) \\
    Permeability scaling factor & \(10^{-1}\) \\
    Noise level & 10\% \\
    \hline
\end{tabular}
\label{tab:param_NN}
\end{table}
To train the model, we first generate a full dataset of sample pairs \((\boldsymbol{k}^{(j)}, \boldsymbol{p}^{(j)})\). During each training iteration, 500 samples are selected randomly from this dataset using a permutation-based sampling strategy so that all available samples are used throughout training. For validation, a fixed set of 200 samples is retained and used consistently across all iterations. This setup allows us to assess convergence and generalization while maintaining a randomized training process.

\subsection{Quantification of inference errors}
To evaluate inference accuracy and define the training objectives, we introduce loss functions that measure errors in the inferred permeability parameterization and, for the physics-informed case, in the corresponding pressure response. For each sample \(j\), we aim that
\[
\widehat{\boldsymbol{k}}^{(j)} = \mathcal{G}_{\Theta}\!\left(\boldsymbol{p}^{(j)}\right).
\]

Here, \(\widehat{\boldsymbol{k}}^{(j)}\) denotes the coefficient vector in the truncated KL representation. In the present work, the supervised targets are taken to be these retained KL coefficients directly, rather than eigenvalue-scaled amplitudes. This choice is motivated by the fact that the KL truncation already restricts the representation to the dominant modes of the permeability field. 

For the data-driven model, the batch loss is defined as the mean squared error between the inferred and reference KL coefficient vectors, where $\|\cdot\|$ denotes the Euclidean vector norm: 
\begin{equation}
\mathcal{L}_{\mathrm{coef}}(\Theta)
=
\frac{1}{N_s}
\sum_{j=1}^{N_s}
\frac{1}{N_{\mathrm{KL}}}
\left\|
\widehat{\boldsymbol{k}}^{(j)}-\boldsymbol{k}^{(j)}
\right\|^2.
\label{eq:Lcoef}
\end{equation}

For the physics-informed model, the inferred coefficient vector is additionally passed through the differentiable forward model \(\mathcal{F}\) to generate the corresponding pressure response at the monitoring locations. The pressure mismatch loss is defined by
\begin{equation}
\mathcal{L}_{\mathrm{pres}}(\Theta)
=
\frac{1}{N_s}
\sum_{j=1}^{N_s}
\frac{1}{N_{\mathrm{obs}}}
\left\|
\mathcal{F}\!\left(\widehat{\boldsymbol{k}}^{(j)}\right)-\boldsymbol{p}^{(j)}
\right\|^2.
\label{eq:Lpres}
\end{equation}

The total batch loss for the physics-informed model is defined as
\begin{equation}
\mathcal{L}_{\mathrm{PI}}(\Theta)
=
\mathcal{L}_{\mathrm{pres}}(\Theta)
+
\alpha_{\mathrm{coef}}\,\mathcal{L}_{\mathrm{coef}}(\Theta),
\label{eq:PIobj}
\end{equation}
where \(\alpha_{\mathrm{coef}} > 0\) is a user-defined weighting parameter. In practice, \(\alpha_{\mathrm{coef}}\) is chosen empirically so that the coefficient and pressure loss terms have comparable magnitude during training. Unless otherwise noted, in this study \(\alpha_{\mathrm{coef}}=0.1\) is used across all physics-informed training cases.

During training, the objective is to minimize the average batch loss over all \(N_b\) batches in an iteration:
\begin{equation}
\mathcal{J}(\Theta)
=
\frac{1}{N_b}
\sum_{i=1}^{N_b}
\mathcal{L}^{(i)}(\Theta).
\end{equation}
For the data-driven model, \(\mathcal{L}^{(i)}(\Theta)=\mathcal{L}_{\mathrm{coef}}^{(i)}(\Theta)\), whereas for the physics-informed model, \(\mathcal{L}^{(i)}(\Theta)=\mathcal{L}_{\mathrm{PI}}^{(i)}(\Theta)\).

\subsection{Reservoir characterization methods}
With both the full physics simulator and the neural network model in place, we established the training methods for reservoir characterization models.  We considered two distinct approaches: a data-driven model and a differentiable simulator-embedded physics-informed method. The data-driven model serves as a benchmark, enabling us to highlight the limitations of learning solely from data, while the physics-informed approach integrates domain knowledge to enhance predictive capability. Both methods are detailed in the following sections.

\subsubsection{Purely data-driven training method}
\begin{figure}[h!]
\centering
\includegraphics[width=0.55\linewidth]{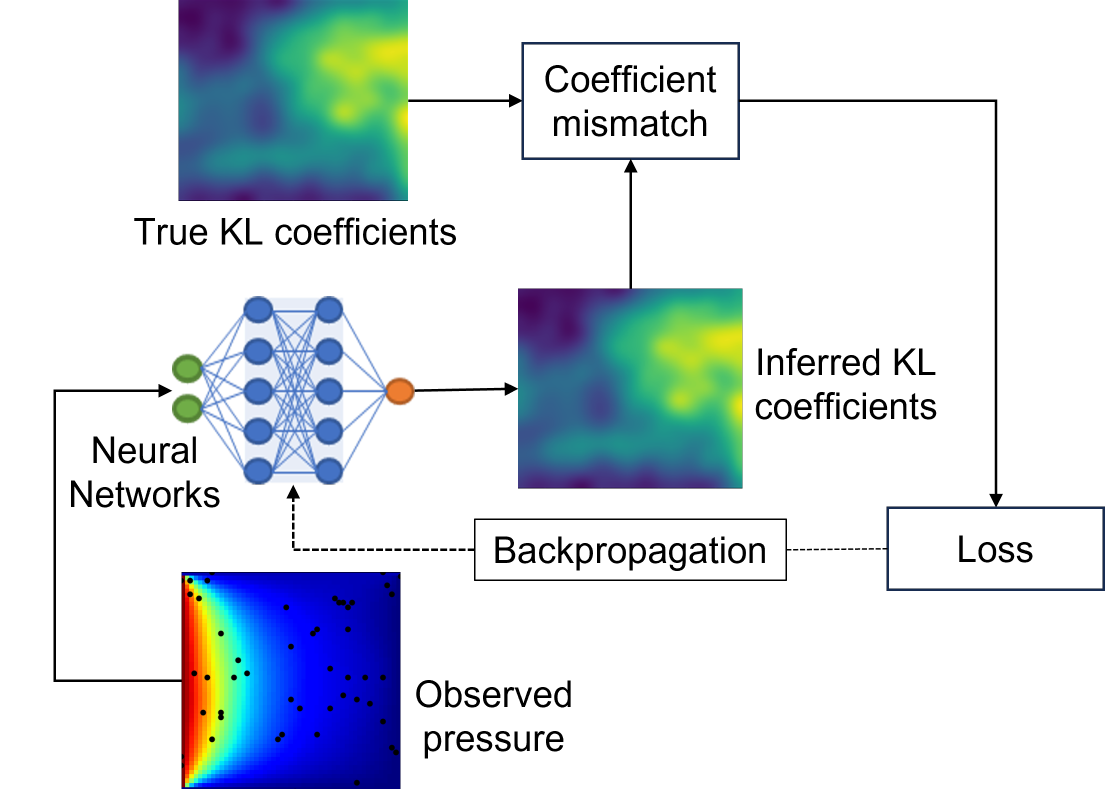}
\caption{Training method for the data-driven model, which shows the method for NN model training using only precomputed permeability-pressure pairs without consideration of any physical constraint. 
}
\label{fig:workflowDataOnly}
\end{figure}

In the data-driven approach, the neural network is trained on precomputed permeability–pressure pairs. Figure \ref{fig:workflowDataOnly} illustrates the training method for this model. The input consists of precomputed pressure values at the observation locations, and the network outputs the corresponding permeability field.  Because no simulation is performed during training, the model does not incorporate pressure mismatch into its objectives and the training loss is defined solely by the mismatch between the inferred and reference permeability coefficients. This design has low computational cost and enables rapid learning. However, its inference accuracy ultimately depends on how well the training data distribution reflects the underlying reservoir physics and on the noise level in the data. 

This model is expected to be well-suited for reservoirs with simpler permeability distribution. However, for complex reservoir heterogeneity, often matching only the coefficient fails to capture the true pressure response within the system. For the validation of this method, we calculate both permeability loss and pressure loss by running the simulator on the inferred permeability field to assess the model's ability to generalize the physical behavior.

\subsubsection{Physics-informed method}
\begin{figure}[h!]
\centering
\includegraphics[width=0.70\linewidth]{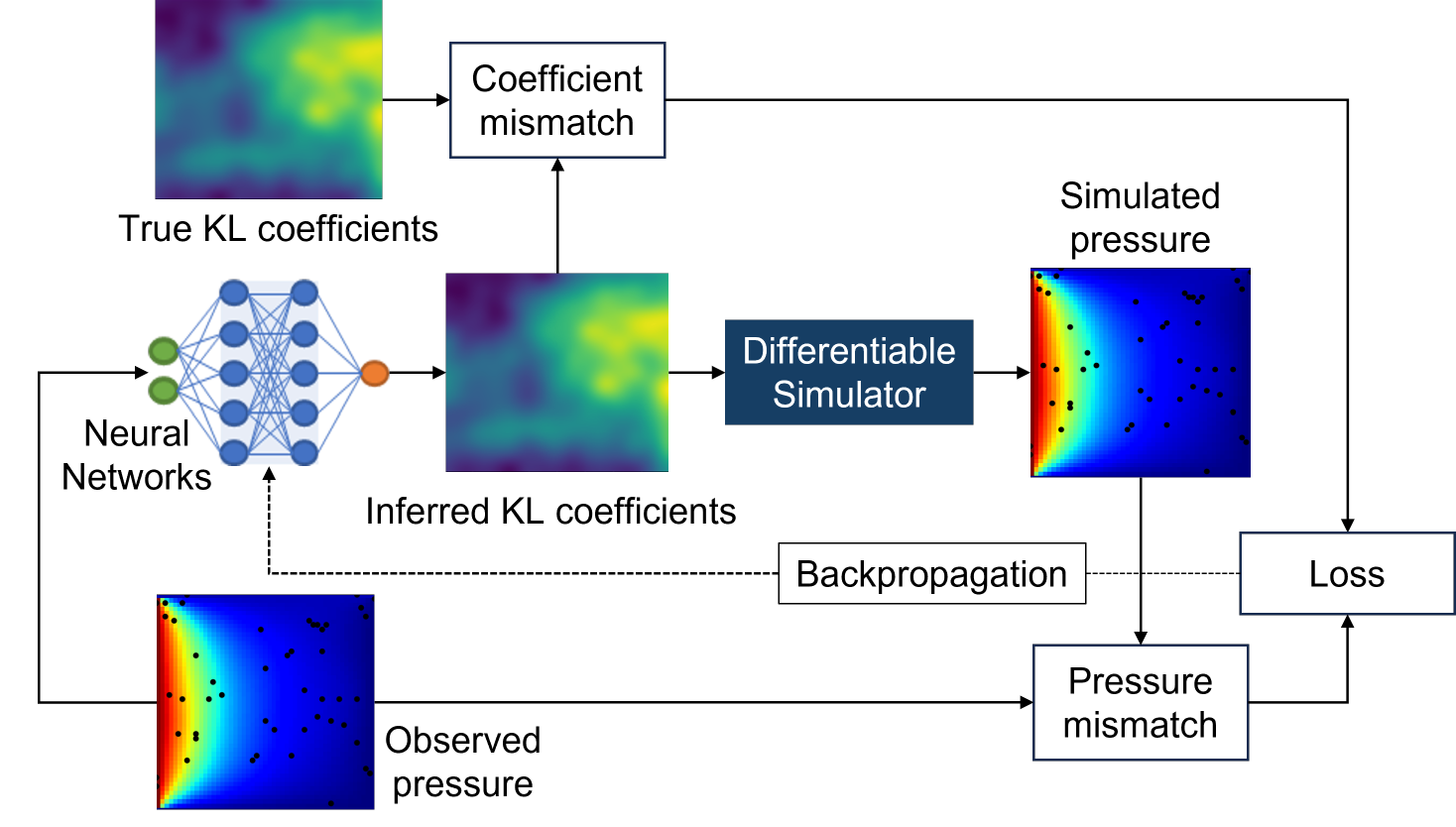}
\caption{Method for the physics-informed learning, where the NN model learn from both pressure and permeability miss match having a differentiable simulator within the training loop.}
\label{fig:workflowSim}
\end{figure}

We train the physics-informed model using both permeability and pressure mismatch. As shown in Figure \ref{fig:workflowSim}, each forward pass infers a permeability field based on observed pressure values, which is then passed through a differentiable simulator to compute the corresponding pressure profiles. We refer to these as the inferred pressures. The total training loss is defined as the sum of the mismatch between inferred and true pressures and the permeability coefficients mismatch. Gradients of the pressure loss with respect to the inferred permeability are obtained via automatic differentiation through the simulator and are subsequently back-propagated through the neural network. This design ensures that the model learns permeability patterns that not only align with the reference coefficients but also reproduce observed pressure behavior.
Embedding the simulator in the training loop enforces physical consistency and reduces the risk of overfitting to patterns that match permeability coefficients but violate flow physics. By including the pressure loss in the objective, the model is explicitly penalized for permeability inferences that fail to reproduce observed pressure behavior. This integration helps the model extrapolate better to unseen well conditions and production scenarios, and makes it more robust to distribution shifts between the training data and the true reservoir. However, the trade‐off is substantially increased cost per training step due to repeated simulator calls during training. 

\section{Results}\label{sec:results}
In this section, we compare the developed methods in predicting the pressure and permeability of subsurface systems. We first show a base case scenario, where we discuss the learning efficiency of the models and evaluate the model's performance on predicting permeability and pressure response using test cases. At the end of this section, we use eight different training scenarios to show the application of the methods in changing reservoir settings and data conditions.

\subsection{Training and validation of the proposed methods}
\begin{figure}[h!]
    \centering
    \begin{subfigure}[b]{0.49\textwidth}
        \includegraphics[width=\textwidth]{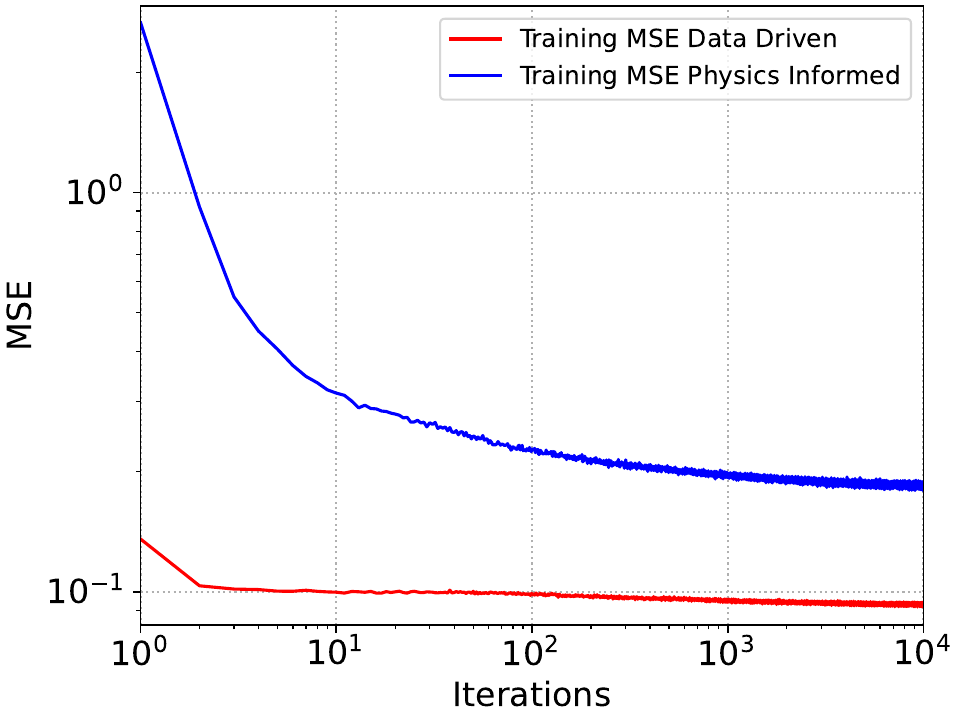}
        \caption{}
        \label{fig:train_err_basecase}
    \end{subfigure}
    \hfill 
    \begin{subfigure}[b]{0.49\textwidth}
        \includegraphics[width=\textwidth]{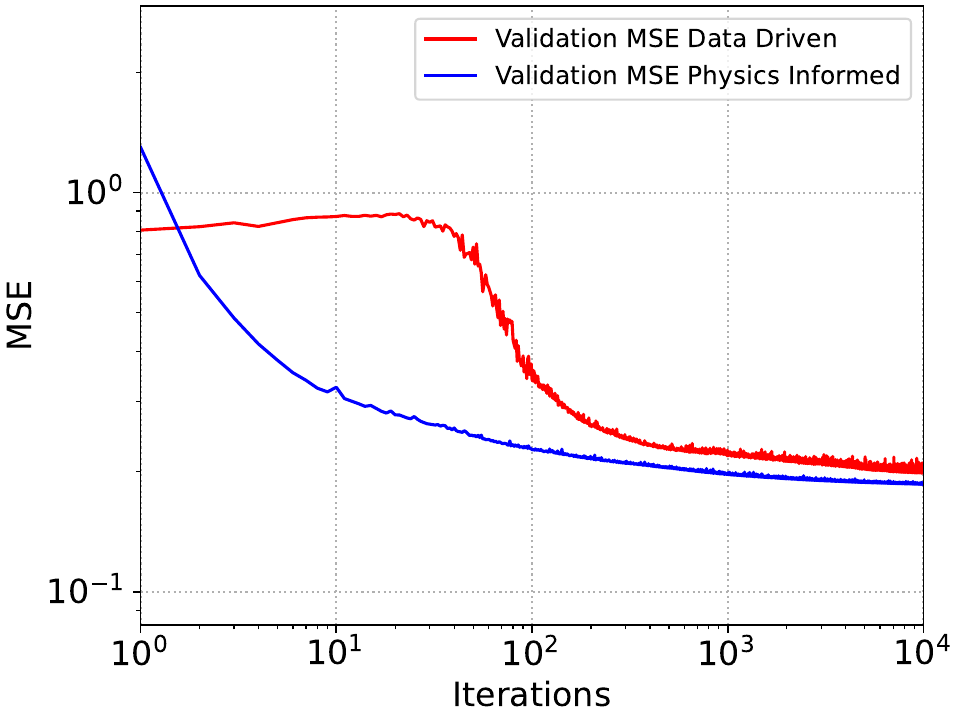}
        \caption{}
        \label{fig:val_err_basecase}
    \end{subfigure}
    \caption{Training and and validation errors for the physics-informed and data-driven model in base case scenario. (a) Training errors, (b) validation errors. The lower training errors for the data-driven model is the result of having only the permeability error in the loss calculation.
    } 
    \label{fig:train_val_err_basecase}
\end{figure}

For the base case, we use 50,000 permeability-pressure pairs for the training and 200 pairs for validation. We set the objective of the NN model to infer the 200 KL coefficients based on the pressure values at 200 observation points. The physics parameters used in this study are listed in Table \ref{tab:param_physics_base}. 
\begin{table}[!h]
\caption{Parameters used in the physics-based model.}
\centering
\begin{tabular}{l c c}
    \hline
        Parameters & Value &  Unit\\
   \hline
    Domain size & 200 $\times$ 200 & m\\
    Number of grid blocks & 51 $\times$ 51 &-\\
   Number of observation nodes & 200 &-\\
    Left-right  dirichlet pressure & [10 0] & MPa \\
    Top-bottom dirichlet pressure  & [0.5 0] & MPa \\
    Variogram (Covariance) type & Matern, exponential & -\\
    Correlation length  & 100  &  m \\
    Sampling Method  & Karhunen–Loève (KL) & -  \\
    Number of KL modes  & 200 & -\\
    \hline
\end{tabular}
\label{tab:param_physics_base}
\end{table}
Using the training procedures shown in Figures \ref{fig:workflowDataOnly} and \ref{fig:workflowSim}, we train both the data-driven and physics-informed models. The resulting training and validation errors are shown in Figure \ref{fig:train_val_err_basecase}. In Figure \ref{fig:train_err_basecase}, it can be seen that both models display stable convergence: the training errors decrease rapidly at early iterations and then levels off. The data-driven model consistently achieves lower training error because it minimizes only the KL-coefficient mismatch, whereas the physics-informed model must additionally satisfy the simulator-based pressure constraint. The difference is more pronounced in the validation graphs shown in Figure \ref{fig:val_err_basecase}. There, the physics-informed model reduces validation error rapidly and remains below the data-driven model for most of training, while the data-driven model converges more slowly. These results suggest that incorporating the physics-based pressure constraint leads to better generalization.

\begin{figure}[h!]
    \centering
    \begin{subfigure}[b]{0.49\textwidth}
        \includegraphics[width=\textwidth]{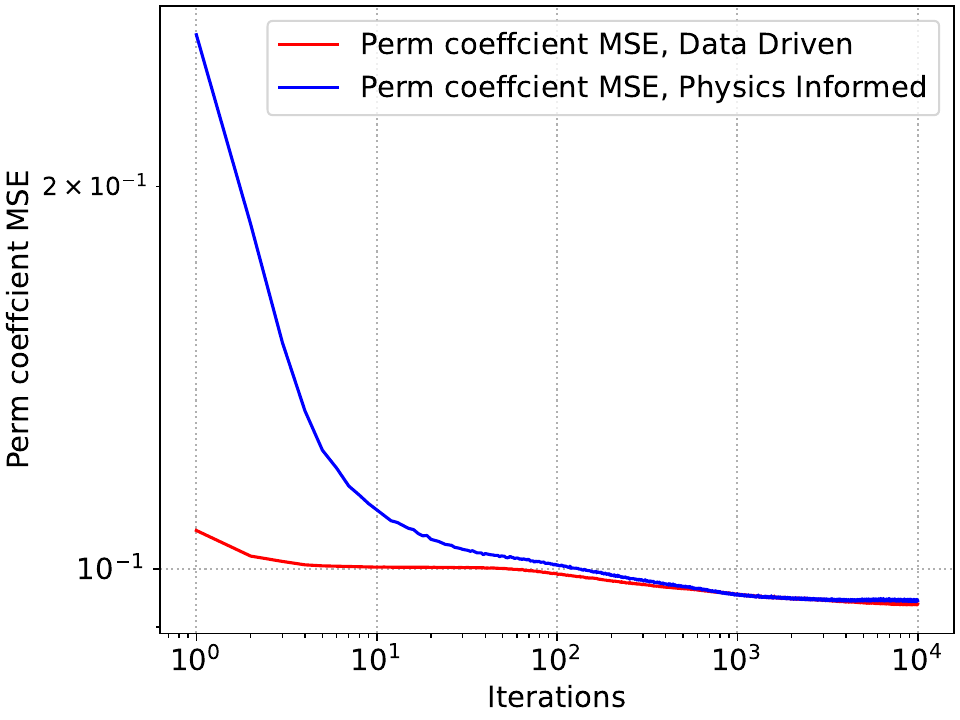}
        \caption{}
        \label{fig:val_err_perm_basecase}
    \end{subfigure}
    \hfill 
    \begin{subfigure}[b]{0.49\textwidth}
        \includegraphics[width=\textwidth]{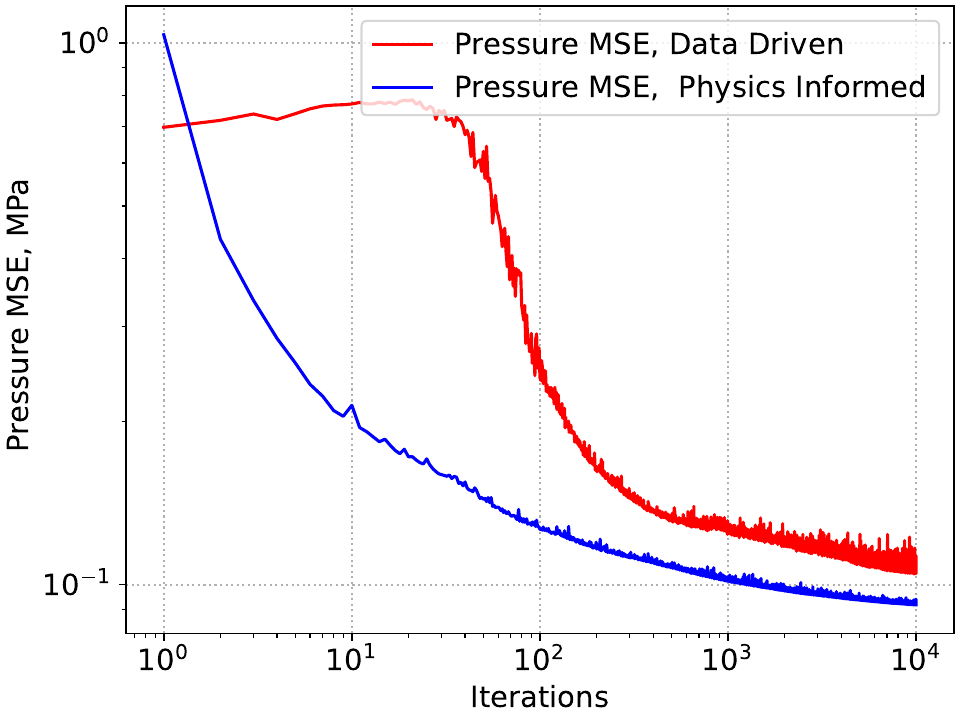}
        \caption{}
        \label{fig:val_err_pres_basecase}
    \end{subfigure}
    \caption{Pressure and permeability errors from the validation data. (a) Permeability errors is approximately the same for the data-driven and physics-informed models towards the end of the training (b) pressure error is lower in the physics-informed approach as it accounts for the pressure mismatch in the loss function.}
    \label{fig:val_pres_and_perm_err}
\end{figure}

To further examine the validation errors, we separate the KL-coefficient mismatch and the pressure mismatch and plot them individually in Figure \ref{fig:val_pres_and_perm_err}. Figure \ref{fig:val_err_perm_basecase} shows that the KL-coefficient mismatch is approximately the same for the data-driven and physics-informed models, indicating that adding the pressure-mismatch term does not significantly degrade the network’s ability to infer the permeability field. However, comparable coefficient errors do not necessarily translate into capturing the complex flow physics, such as pressure behavior. As shown in Figure \ref{fig:val_err_pres_basecase}, the physics-informed model produces substantially lower pressure error than the data-driven model. This demonstrates the advantage of incorporating physics into the learning process for inverse modeling, leading to more physically consistent permeability inferences.
 
Next, we use the trained models to predict permeability and pressure for a random test case that was not part of the training or validation datasets. A randomly selected permeability sample is used as the true permeability, and we simulate the domain with the full-physics solver using the parameters listed in Table \ref{tab:param_physics_base}. The true pressure response at the monitoring locations is then recorded. 

\begin{figure}[h!]
    \centering
    \begin{subfigure}[b]{0.99\textwidth}
        \includegraphics[width=\textwidth]{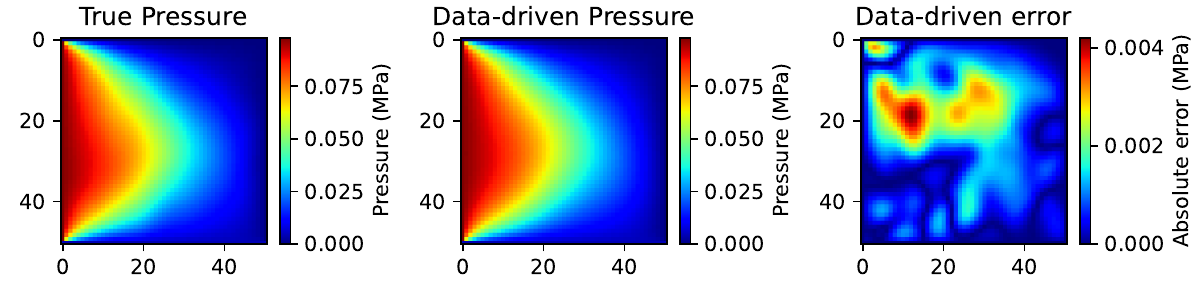}
        \caption{}
        \label{fig:Application_pressure_test_perm}
    \end{subfigure}
    \hfill 
    \begin{subfigure}[b]{0.99\textwidth}
        \includegraphics[width=\textwidth]{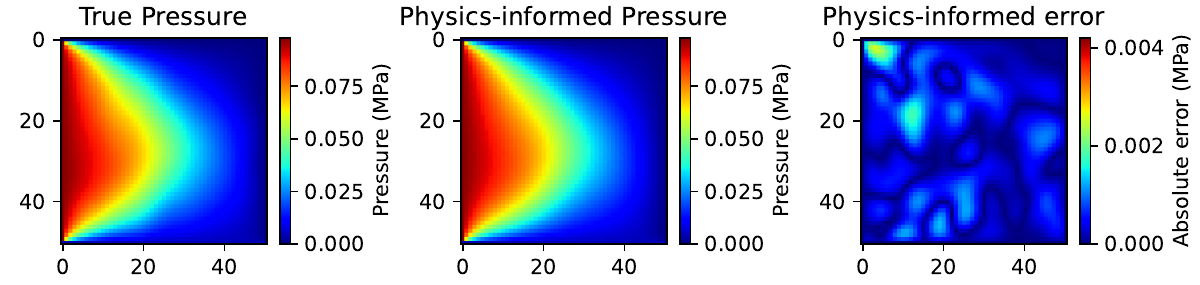}
        \caption{}
        \label{fig:Application_pressure_test_contour}
    \end{subfigure}
    \caption{Evaluation of the trained model. (a) Pressure inference error from the data-driven model; (b) pressure inference error from the physics-informed model. The data-driven model produces significantly higher error value in matching the true pressure field.
    }
    \label{fig:pres_comparison_basecase}
\end{figure}

Figure \ref{fig:pres_comparison_basecase} shows the pressure fields obtained from the inferred permeability fields. The difference between the two models is more pronounced in pressure than in permeability. The rightmost plots show that the data-driven model yields larger pressure errors than the physics-informed model. In this test case, the maximum pressure error is 0.004 MPa for the data-driven model, compared with 0.002 MPa for the physics-informed model. This indicates that the physics-informed model reproduces the pressure response more accurately. 

\begin{figure}[h!]
    \centering
    \begin{subfigure}[b]{1\textwidth}
        \includegraphics[width=\textwidth]{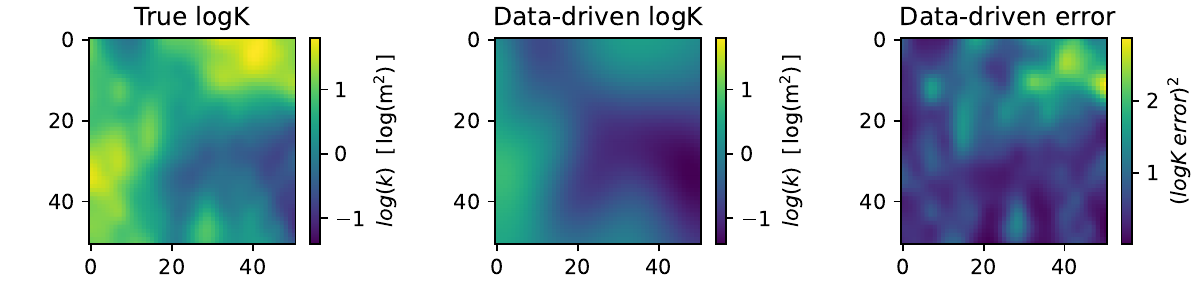}
        \caption{}
        \label{fig:Application_pressure_test_perm}
    \end{subfigure}
    \hfill 
        \begin{subfigure}[b]{1\textwidth}
        \includegraphics[width=\textwidth]{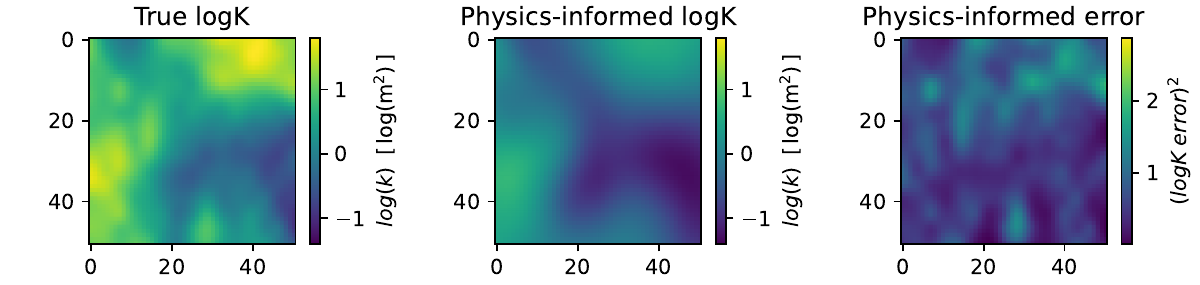}
        \caption{}
        \label{fig:Application_pressure_test_contour}
    \end{subfigure}
    \caption{Evaluation of the trained model. (a) Permeability inference error from the data-driven model; (b) permeability inference error from the physics-informed model. The error differences are insignificant, however, the physics-informed model provides a slightly better permeability inference.}
    \label{fig:perm_comparison_basecase}
\end{figure}

Figure \ref{fig:perm_comparison_basecase} compares the inferred permeability fields for the two models. The top row shows the reference permeability, the permeability inferred by the data-driven model, and the corresponding error field, while the bottom row shows the same quantities for the physics-informed model. The two models produce broadly similar permeability inferences, although the physics-informed model appears to agree slightly better with the reference field. Taken together, these results show that visually similar permeability inferences do not necessarily lead to equally accurate pressure predictions.

After presenting a visual comparison from one test case, we now focus on a systematic comparison in which we 
study inferences on several different test data samples. We first generate 1,000 random permeability samples based on the setup and parameters discussed above. For each sample, we compute the corresponding pressure response using the full-physics simulator and save the permeability–pressure pairs for model evaluation. The observed pressures are then used as input into the trained models, which predict the permeability fields. For each test sample, the pressure error is quantified by the root-mean-square error (RMSE) between the inferred and reference pressures at the monitoring locations. The permeability error is quantified by the relative $L_2$-norm error between the inferred and true log-permeability fields.

\begin{figure}[h!]
    \centering
    \begin{subfigure}[b]{0.33\textwidth}
        \includegraphics[width=\textwidth]{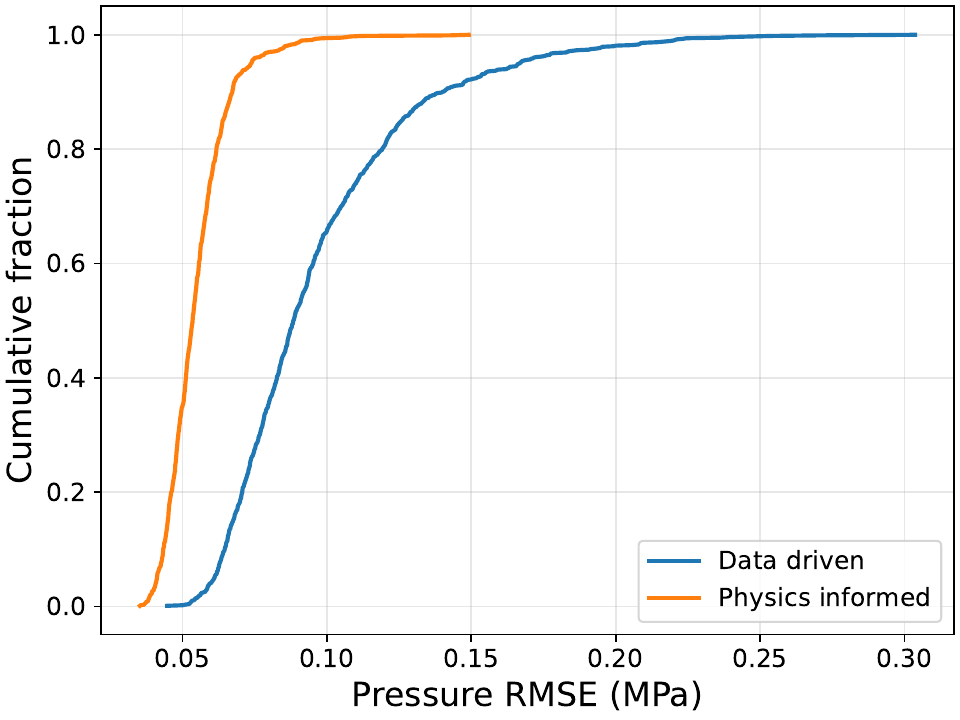}
        \caption{}
        \label{fig:ecdfbhp_llll}
    \end{subfigure}
    \hfill 
    \begin{subfigure}[b]{0.33\textwidth}
        \includegraphics[width=\textwidth]{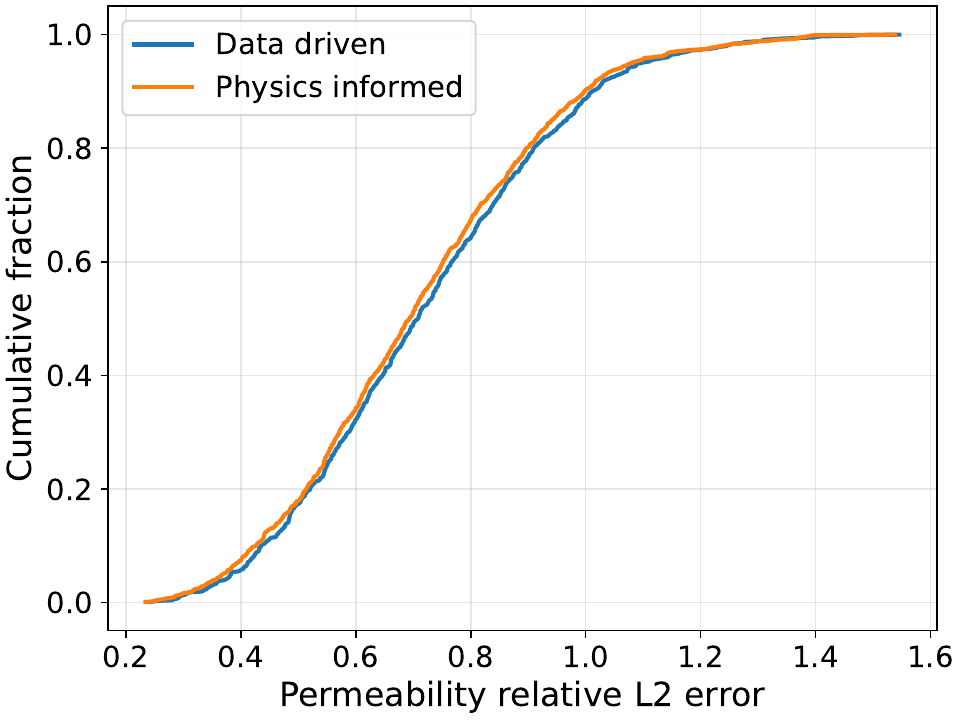}
        \caption{}
        \label{fig:ecdfperm_llll}
    \end{subfigure}
    \hfill
    \begin{subfigure}[b]{0.33\textwidth}
        \includegraphics[width=\textwidth]{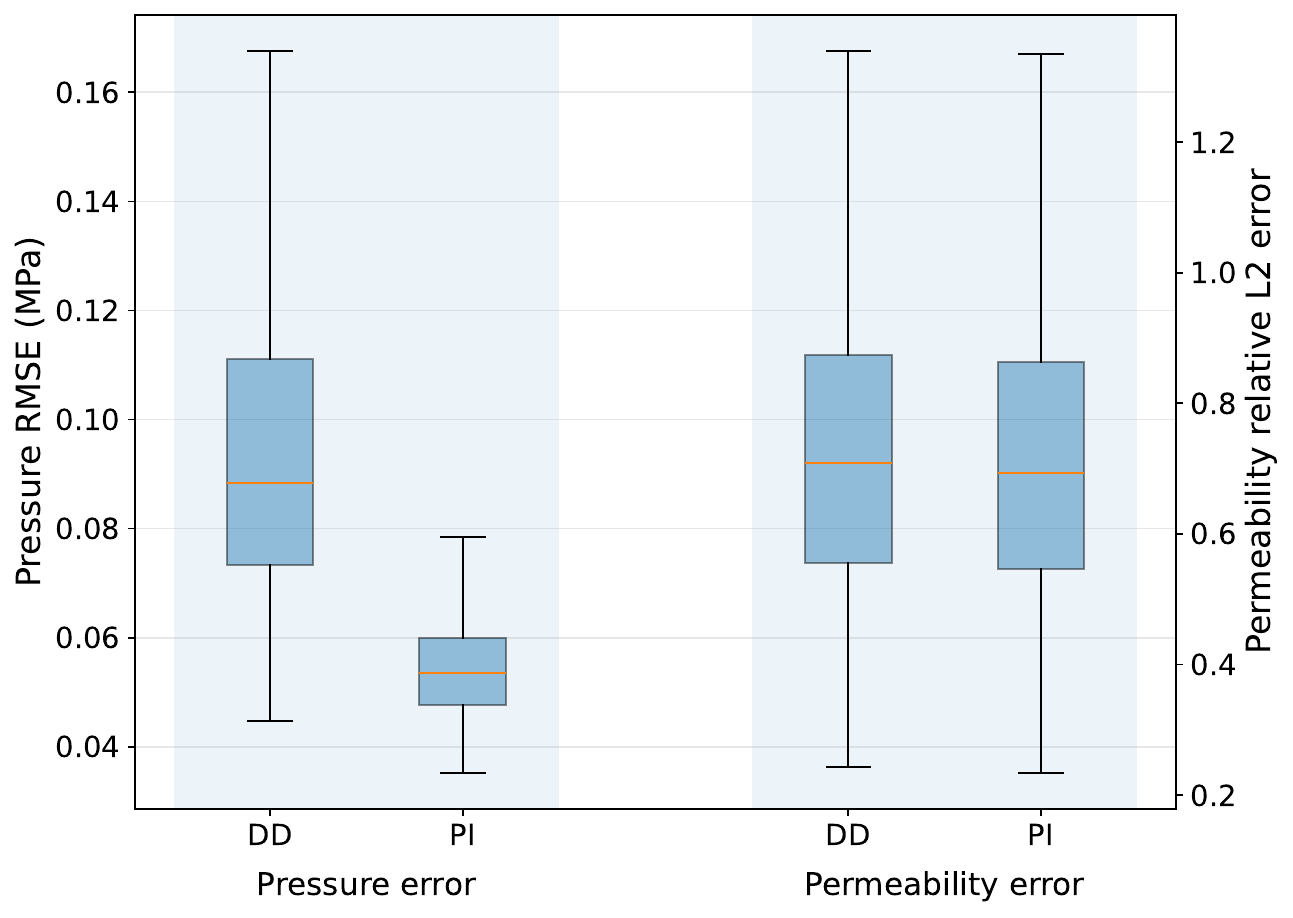}
        \caption{}
        \label{fig:box_lll}
    \end{subfigure}
    \hfill 
        \begin{subfigure}[b]{0.33\textwidth}
        \includegraphics[width=\textwidth]{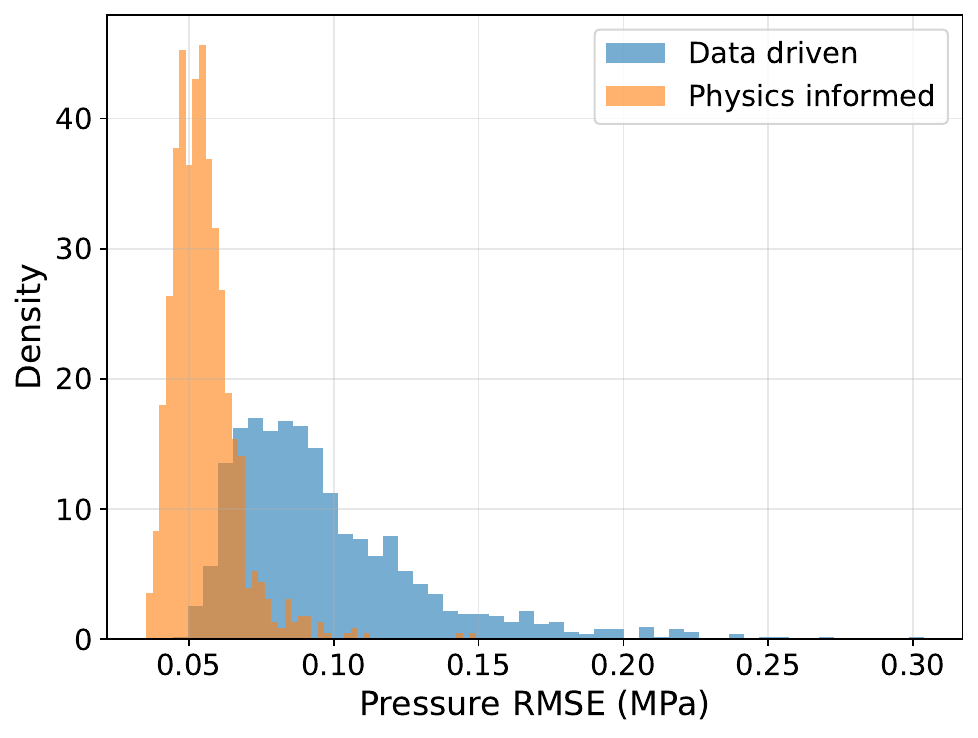}
        \caption{}
        \label{fig:histnhp_lll}
    \end{subfigure}
    \hfill 
        \begin{subfigure}[b]{0.33\textwidth}
        \includegraphics[width=\textwidth]{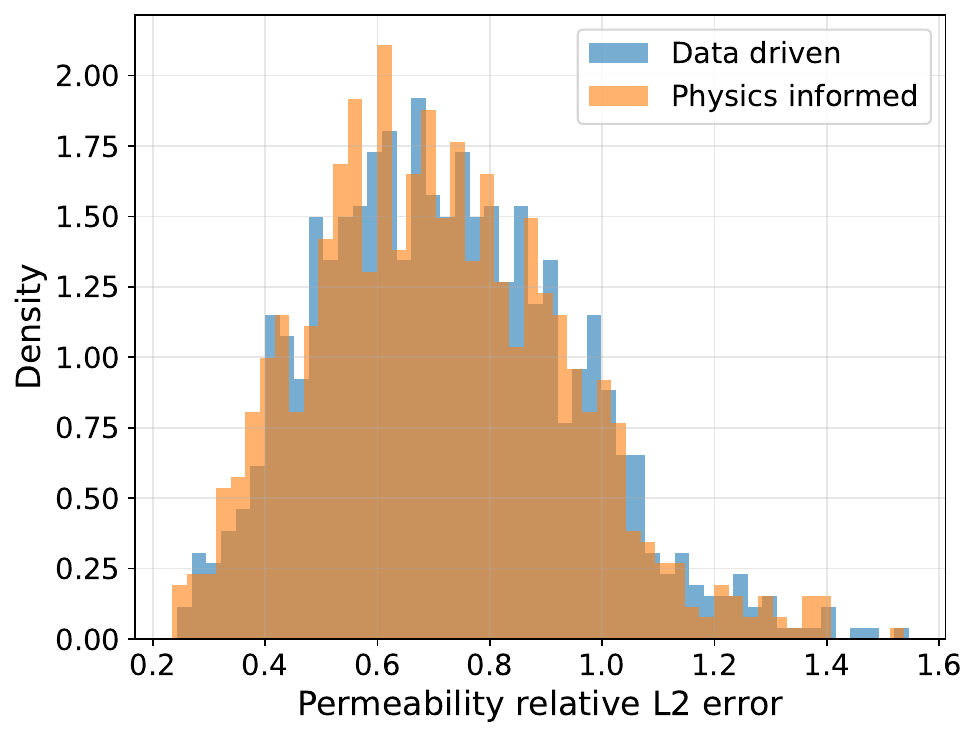}
        \caption{}
        \label{fig:histperm_lll}
    \end{subfigure}
    \hfill 
    \begin{subfigure}[b]{0.33\textwidth}
        \includegraphics[width=\textwidth]{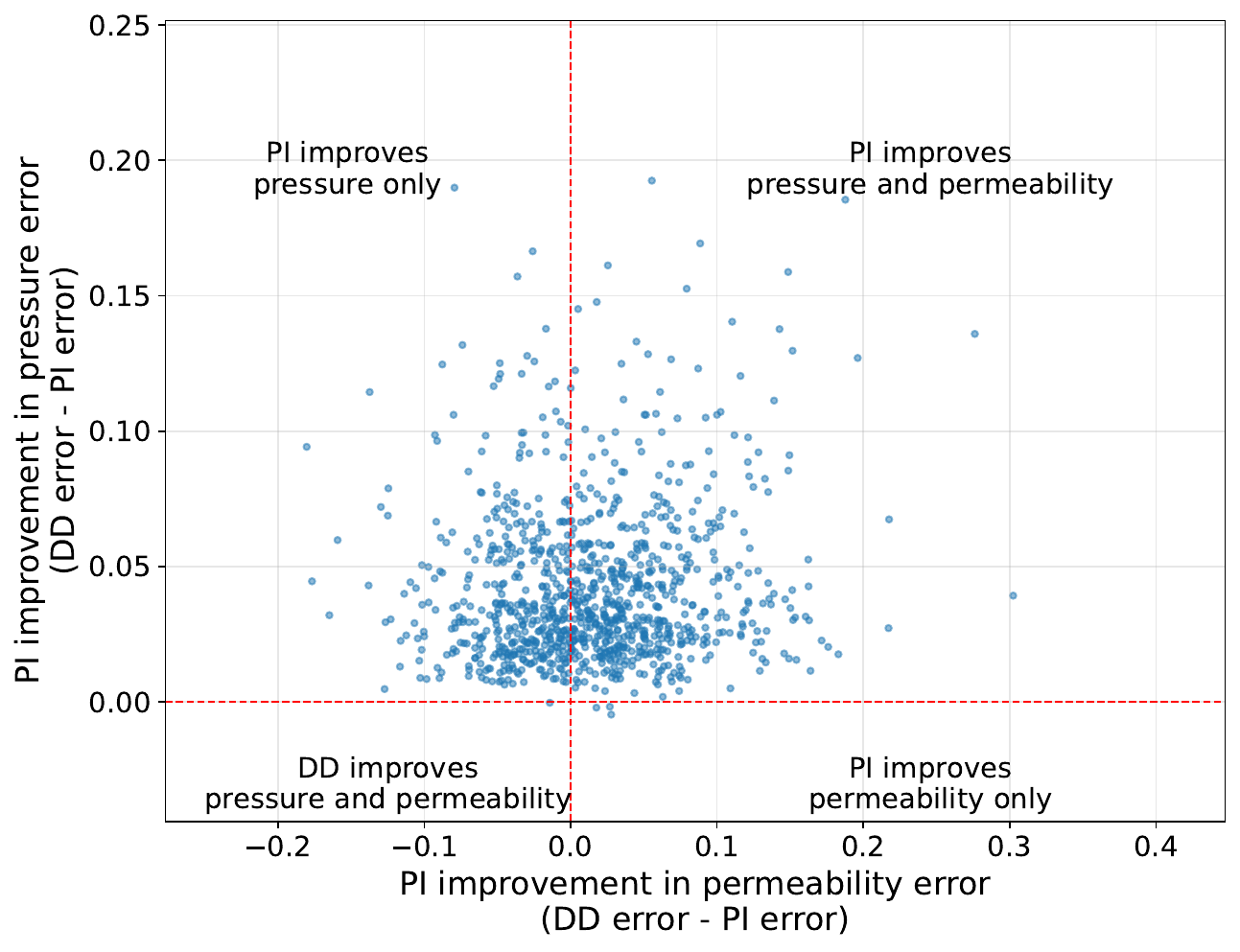}
        \caption{}
        \label{fig:scatter_lll}
    \end{subfigure}
    \caption{Statistical distribution of inference error. (a) Empirical cumulative distribution function (ECDF) for pressure inference errors. (b) ECDF for permeability inference errors. (c) Histogram of pressure errors. (d) Histogram of permeability errors. (e) Box plot comparison of model performance (Left axis: Pressure MSE; Right axis: Permeability $L_2$ Error). (f) Relative error scatter between physics-informed and data-driven inference. These plots confirm that the physics-informed model performs more accurately in pressure inference while maintaining comparable accuracy in permeability inference. }
    \label{fig:stat_lll}
\end{figure}

The results are presented in Figure \ref{fig:stat_lll}, where we compare the trained model's inference ability. Figure \ref{fig:ecdfbhp_llll} shows the empirical cumulative distribution function (ECDF) of pressure errors for both the data-driven and the physics-informed cases. The ECDF plot indicates which fraction of the data that falls within a given error limit. We observe that, for 80\% of the samples, the errors in the physics-informed case are nearly half of those in the data-driven case. Figure \ref{fig:ecdfperm_llll} shows the ECDF plot for the permeability error. In this plot, we see that the physics-informed model produces slightly lower error values: 80\% of the inferences have errors less than 0.89 m$^2$, compared to 0.90 m$^2$ for the data-driven model.

Figure \ref{fig:box_lll} shows the box plot of the errors. This figure is useful for quantifying and visually representing the error range and mean error values of the models. We use two different axes in this plot: the left axis shows the mean squared pressure error, while the right axis shows the relative L2-norm of the permeability error. The two boxes on the left represent the pressure errors for the data-driven (DD) and physics-informed (PI) cases, respectively, while the two boxes on the right correspond to the permeability errors. The central line within each box marks the median. The box spans the interquartile range (Q1–Q3); the whiskers extend to the most extreme points within 1.5 times the interquartile range, and outliers are omitted to emphasize bulk behavior.
From the plots, we can see that the pressure errors in the data-driven case range approximately between 0.04 and 0.16 MPa, with a median value of 0.09 MPa. However, the range for the physics-informed model is between 0.035 and 0.08 MPa. The two plots on the right show that the permeability error medians are close -- 0.71 and 0.70 m$^2$ for the data-driven and physics-informed models, respectively. The error range is similar for both cases.

Figure \ref{fig:histnhp_lll} shows the histogram of pressure errors. From the distribution, we can see that the errors from the data-driven model are more widely spread than those from the physics-informed model. The error range for the physics-informed model is between 0.01 MPa and 0.15 MPa, while for the data-driven model it ranges from 0.01 MPa to 0.32 MPa. Figure \ref{fig:histperm_lll} shows a similar histogram for the permeability errors. The permeability distributions are not significantly different between the two models, as the relative $L_2$ norm errors from both models overlap. This indicates that both models have comparable capability to predict permeability in the given training scenario.

Figure \ref{fig:scatter_lll} compares the sample-wise improvement of the physics-informed model relative to the data-driven model in both pressure and permeability inference. Each point represents a test sample. The $x$-axis shows the physics-informed improvement in permeability error, defined as the data-driven permeability error minus the physics-informed permeability error, while the $y$-axis shows the corresponding improvement in pressure error. The dashed red lines at zero divide the plot into four quadrants. Points in the upper-right quadrant indicate samples for which the physics-informed model improves both pressure and permeability inference. Points in the upper-left quadrant indicate improvement only in pressure, whereas points in the lower-right quadrant indicate improvement only in permeability. The lower-left quadrant corresponds to samples for which the data-driven model performs better in both quantities. The distance from the origin reflects the magnitude of the performance difference between the two approaches. From the plot, most samples lie above the horizontal zero line and are distributed nearly symmetrically about the vertical axis, indicating that the physics-informed model consistently improves pressure inference while maintaining broadly comparable permeability inference accuracy.

From the results, we conclude that both the data-driven and physics-informed models perform similarly in permeability inference. However, the physics-informed model shows improved accuracy in pressure inference. Overall, incorporating physics into the learning process had a substantial effect and improved the models' performance in producing physics-consistent inferences.

\subsection{Generalization of the reservoir characterization methods under varying data scenarios}

As shown in the base case, the physics-informed method performs better than the data-driven method in reservoir characterization. We next examine the method's behavior under a broader range of training scenarios relevant to practical subsurface applications. Because reservoir heterogeneity and data availability can vary substantially from case to case, it is important to assess model performance under different observational and geological settings. To this end, we considered eight scenarios defined by three factors: training dataset size, number of observation points, and degree of heterogeneity. The training dataset size was varied between 5,000 and 50,000 permeability--pressure pairs, the number of monitoring points between 50 and 200, and the correlation length of the Gaussian random field between 10~m and 100~m. These combinations define the eight scenarios listed in Table~\ref{tab:train_scenarios}. 

\begin{table}[!h]
\caption{Summary of training scenarios defined by dataset size, number of observations, and correlation length of the permeability field. The assigned case names represent combinations of these three factors.}
\centering
\begin{tabular}{l c c c}
\hline
Training      & Number of        & Correlation & Case \\
data size     & observations     & length      & name \\
\hline
Large (50 k) & Large (200) & Large (100 m) & LLL \\
Large (50 k) & Large (200) & Small (10 m) & LLS \\
Large (50 k) & Small (50) & Large (100 m) & LSL \\
Large (50 k) & Small (50) & Small (10 m) & LSS \\
Small (5 k) & Large (200) & Large (100 m) & SLL \\
Small (5 k) & Large (200) & Small (10 m) & SLS \\
Small (5 k) & Small (50) & Large (100 m) & SSL \\
Small (5 k) & Small (50) & Small (10 m) & SSS \\
\hline
\end{tabular}
\label{tab:train_scenarios}
\end{table}

For each scenario listed in Table \ref{tab:train_scenarios}, we train both physics-informed and data-driven models using the corresponding datasets and evaluate them on 1,000 randomly generated test samples. As summarized in Table \ref{tab:train_scenarios_outcome}, the physics-informed model consistently improves pressure inference relative to the data-driven model, reducing the mean pressure error by 33\% to 64\% across the eight scenarios. By comparison, the improvement in permeability inference is modest. The physics-informed model achieves a lower mean permeability error in seven of the eight cases, with a maximum reduction of 5.6\%. Because the results for the 5,000- and 50,000-sample training sets were qualitatively very similar, we focus the discussion below on the four cases with 5,000 training samples.

\begin{table}[!h]
\caption{Comparison of inference accuracy for the physics-informed  and data-driven models in variable data conditions.}
\centering
\begin{tabular}{l c c c c c c}
\hline
Case  & \% PI wins perm & \% PI wins pres &  PI reduces perm error &  PI reduces pres error\\
\hline
LLL & 59\% &  100\% & -0.8\% & 59.4\%\\
LLS & 97\%& 100\% & 2.4\% & 39.7\%\\
LSL & 56\% & 99\%&3.9\% & 42.3\%\\
LSS & 94\% & 100\% & 3.8\%  & 63.9\%\\
SLL & 59\% & 99\%& 3.4\% & 41.2\%\\
SLS  & 83\% & 100\%& 3.1\%&33.1\%\\
SSL & 65\%& 100\% & 5.6\%&48.8\%\\
SSS  & 95\% & 100\% &4.2\%& 62.2\%\\
\hline
\end{tabular}
\label{tab:train_scenarios_outcome}
\end{table}

\begin{figure}[!h]
\centering
\includegraphics[width=0.99\linewidth]{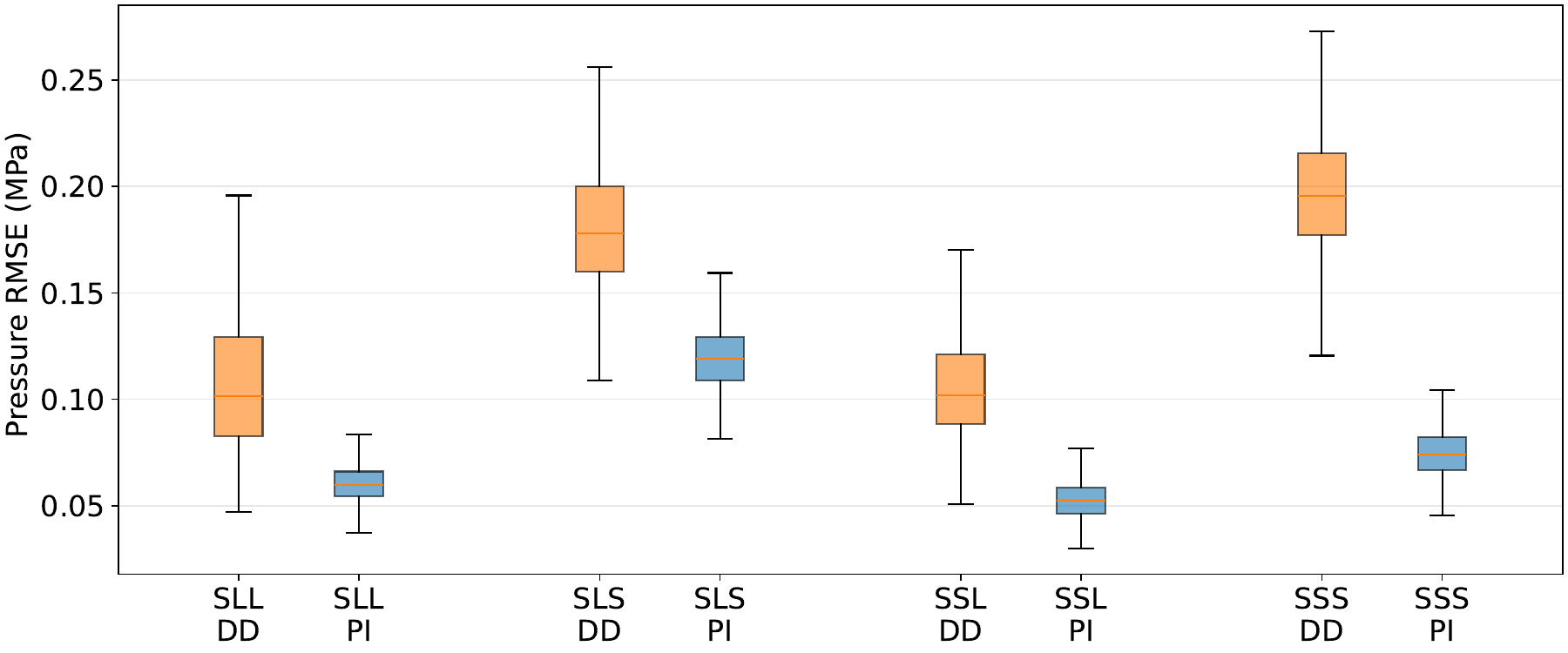}
\caption{Comparison of estimated error in inferred pressure between the physics-informed and data-driven model for the eight data scenarios shows the physics-informed model infers pressure with higher accuracy in every scenario.}
\label{fig:PrescompareOverall}
\end{figure}

\begin{figure}[!h]
\centering
\includegraphics[width=0.99\linewidth]{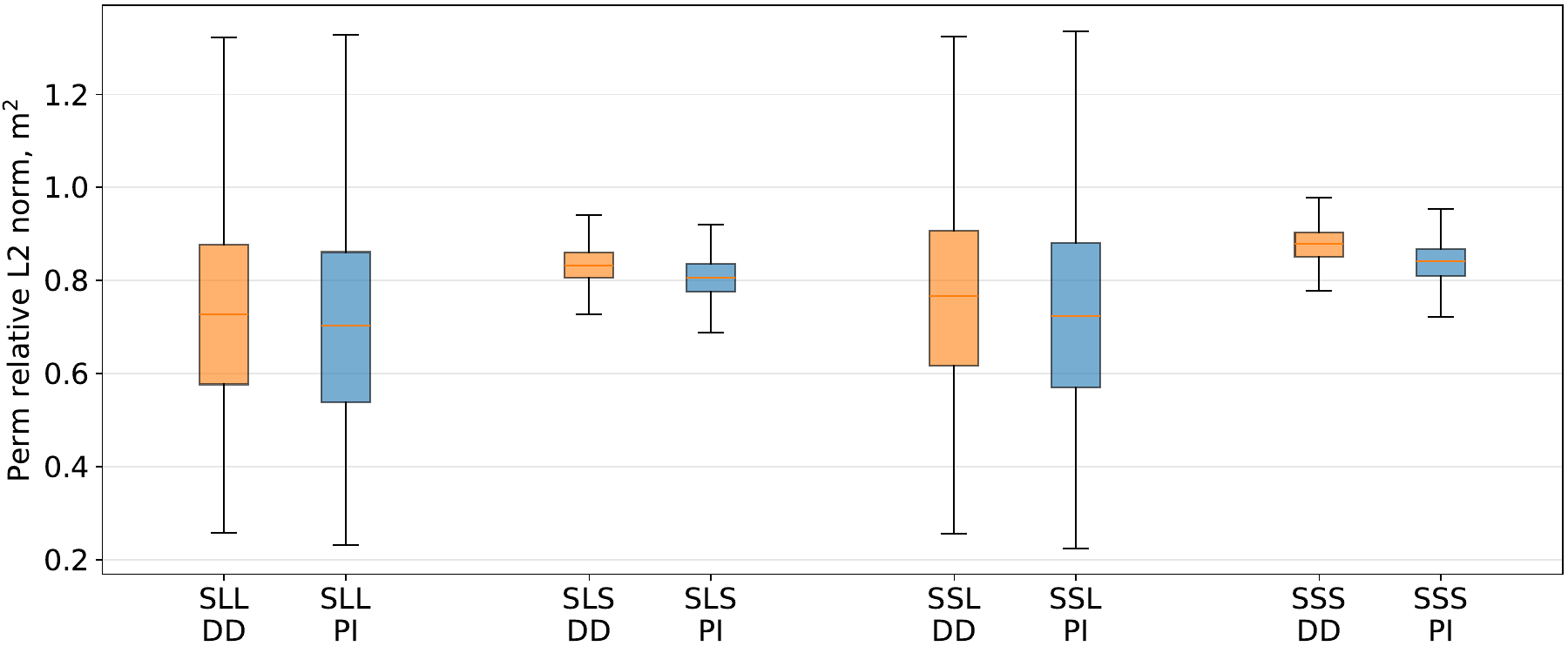}
\caption{Comparison of permeability inference errors between the physics-informed and data-driven models for the eight data scenario shows the physics-informed model perform with similar accuracy in permeability inference despite accounting pressure loss in learning.}
\label{fig:PermcompareOverall}
\end{figure}

In Figure \ref{fig:PrescompareOverall}, we present the pressure errors for each of the training data scenarios using box plots, and Figure \ref{fig:PermcompareOverall} shows the corresponding permeability errors for both models. In the pressure error plots, the physics-informed model has lower errors than the data-driven model in all scenarios. However, in the permeability error plots, we do not observe any significant difference between the physics-informed and data-driven models. The results also show that both the number of observation points and the correlation length have a significant impact on model performance. The difference in pressure error is largest when the number of monitoring points is small and the correlation length is small. A small correlation length enables sharp changes in permeability over small length scales. Under these conditions, the data-driven model struggles to match the observed pressure responses due to limited observational data and the absence of physical constraints.

These results indicate that, despite achieving good permeability matches, data-driven models may fail to capture the true physical response of the system and may therefore mischaracterize the reservoir. They also highlight why physics-informed surrogate models are necessary for reliable reservoir characterization, particularly when the reservoir is highly heterogeneous.

\subsection{Application of the proposed methods for the extreme data scenario}
In reservoir characterization, extreme events may correspond to high pressures caused by specific (and unlikely) permeability field structures. The consequences of such events may include over-pressurization at critical locations or the activation of faults which lead to induced seismicity. Although such events may be rare, they are important and the question arises if the trained map is able to accurately infer permeability in the extreme data regime. Recognizing such events is essential so that appropriate management strategies, mitigation plans, and safety precautions can be implemented in advance. Therefore, it is important to include these samples when training models for reservoir characterization and risk assessment. A model trained only on the bulk distribution will encounter few rare event samples and might not be able to accurately characterize the rare event regime. However, if we extract or generate samples in the extreme event region and include them in the training data along with the bulk samples, the model becomes more effective in predicting system parameters in both the bulk and rare event regimes. 

\begin{figure}[h!]
    \centering
    \begin{subfigure}[b]{0.46\textwidth}
        \includegraphics[width=\textwidth]{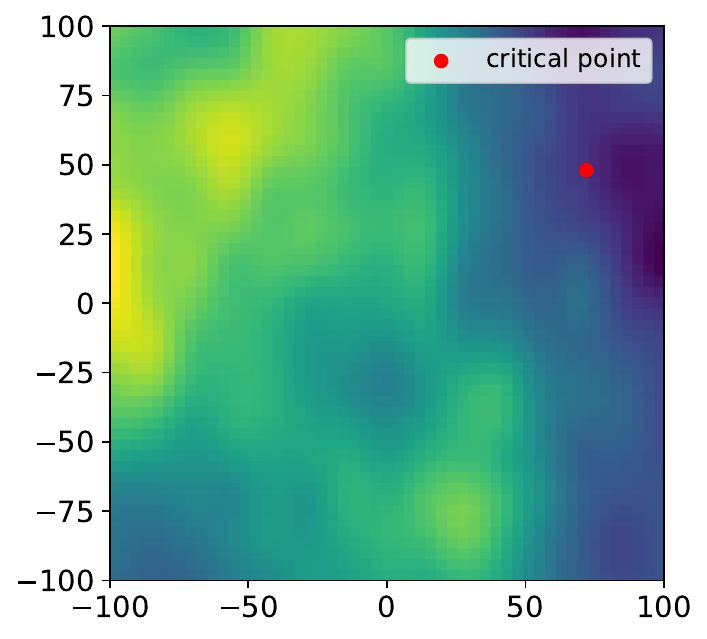}
        \caption{}
        \label{fig:rareevents_crtical_location}
    \end{subfigure}
    \hfill 
    \begin{subfigure}[b]{0.53\textwidth}
        \includegraphics[width=\textwidth]{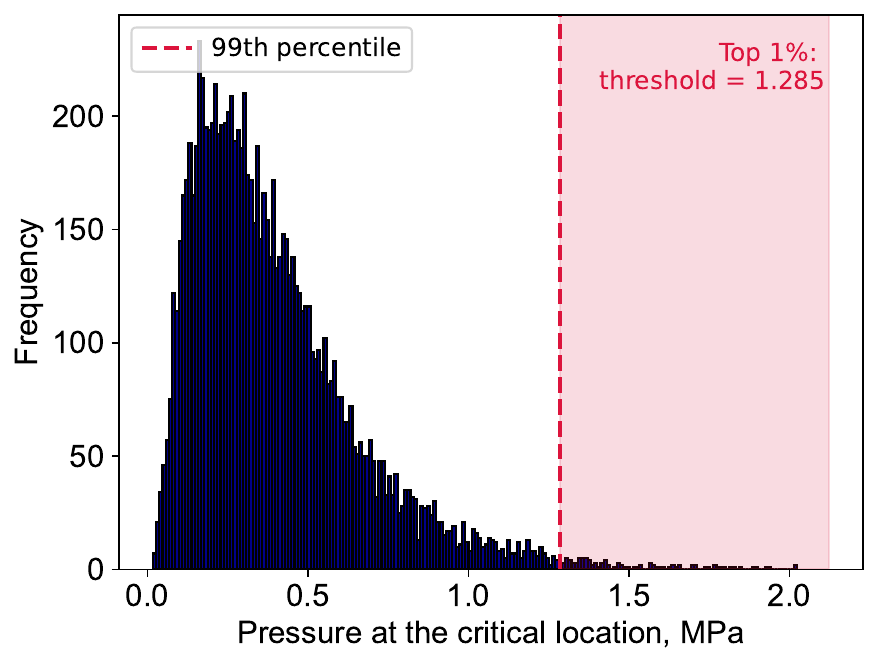}
        \caption{}
        \label{fig:rareevents_pressure_bruteforce}
    \end{subfigure}
    \caption{Generation of rare-event samples. (a) Location of the critical point. (b) Distribution of pressure at the critical location based on 10,000 random permeability samples. Samples which result in pressures above the prescribed threshold value of 1.285 MPa (top 1\%) are selected as rare events.}
    \label{fig:rare_events_bruteforce}
\end{figure}

Exploring extreme events is a challenging task, as they are typically encountered rarely in standard Monte Carlo sampling. Thus, generating tail samples or extreme data can be computationally expensive. A brute-force approach to select extreme data is illustrated in Figure \ref{fig:rare_events_bruteforce}. In this example, our quantity of interest, through which we define extremeness, is the pressure at a critical location. Using the parameters of the base case, we compute the pressure at the critical location for 10,000 randomly generated permeability samples and plot the resulting distribution as a histogram in Figure \ref{fig:rareevents_pressure_bruteforce}. We then select the top 1\% of this distribution as extreme events and record the corresponding permeability and pressure responses to train and evaluate our physics-informed and data-driven models.

Generating extreme samples using this brute-force approach is inefficient, as running 10,000 simulations only yields 100 extreme data, which is not sufficient to train a model or capture a representative distribution of outcomes. To address this, we employ the importance sampling framework developed in \citet{tong2021extreme}. This method utilizes large deviation theory to construct a biasing distribution that focuses sampling efforts specifically on the tail regions of the probability density function. By re-weighting these samples, we can estimate tail probabilities and generate a rich set of extreme events without the computational cost of a large number of standard simulations.

Using this approach, we generate 6,200 rare samples that produce pressure at the critical location that exceeds a prescribed threshold value. We use 5,000 samples as training data and 200 for validation. The remaining 1,000 samples are used to evaluate model performance in the rare event regime.

\begin{figure}[!h]
\centering
\includegraphics[width=0.99\linewidth]{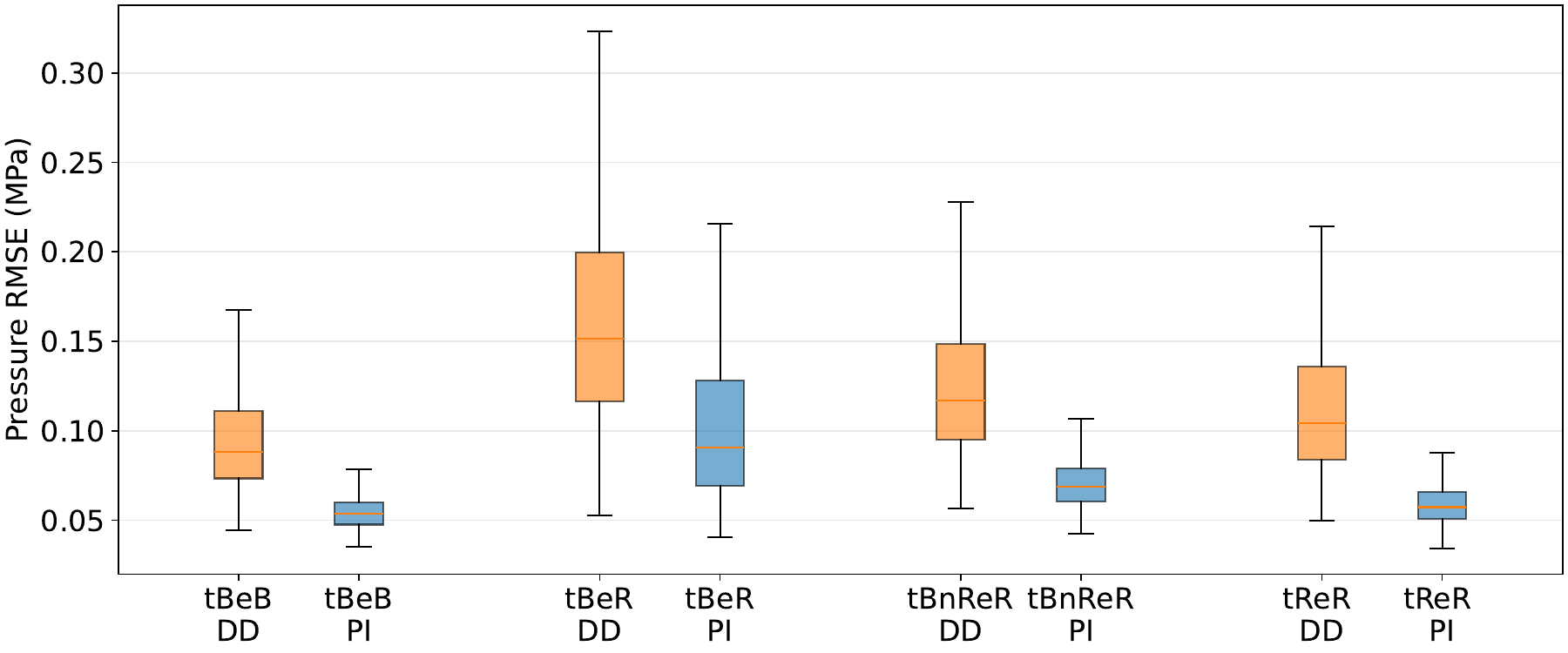}
\caption{Comparison of pressure inference errors in the rare event regime. Models trained only on bulk data (tBeR) show significantly higher error, while incorporating rare event samples into training reduces error for both approaches. The physics-informed model achieves the largest improvement and maintains strong accuracy even when trained solely on rare event data (tReR).}
\label{fig:pres_err_rare_events}
\end{figure}

We study three rare-event cases. First, we test our bulk base model on the 1,000 rare event  samples. We denote this case in Figure \ref{fig:pres_err_rare_events} as tBeR (train on bulk – evaluated on rare). Next, we train a model using the same setup and parameters as in the base case but add 5,000 rare-event samples to the training dataset. After training, we evaluate this model on the rare-events and denote it as tBnReR (train on bulk and rare - evaluated on rare). Finally, we train new data-driven and physics-informed models using only the 5,000 rare event samples and evaluate them on extreme data. We label this scenario as tReR (train on rare - evaluated on rare).

We compare these rare event cases with our bulk base case model, denote tBeB (train on bulk – evaluated on bulk), and show all results in Figure \ref{fig:pres_err_rare_events}. The results indicate that models trained only on bulk data perform poorly when tested on rare-event samples, with significantly higher pressure errors. This implies that bulk-trained models are not well suited for rare event characterization. Adding rare samples to the training dataset substantially reduces the error range for both data-driven and physics-informed models. The improvement is more significant for the physics-informed model. The data-driven model benefits less because the added rare event samples increase variability and complexity in the training dataset, making learning harder without physics constraints. In the final case where we train and evaluate models exclusively on rare-event samples, the physics-informed model achieves nearly the same accuracy as the base-case physics-informed model trained and evaluated on bulk data (tBeB). In contrast, the data-driven model still exhibits significantly higher error. These results highlight the importance of incorporating physics and rare event samples into the training process to ensure robust performance under extreme and safety-critical conditions.

\section{Discussion}\label{sec:discussion}
In this work, we develop and evaluate a physics-informed machine learning framework for reservoir characterization that embeds a fully differentiable single-phase flow simulator within the training loop of a neural network. The network infers a low-dimensional representation of heterogeneous permeability fields in terms of Karhunen–Lo\`eve (KL) coefficients from sparse pressure observations, and the simulator is used to enforce physical consistency through a pressure-mismatch loss term. We compared this physics-informed method with a purely data-driven method that uses the same architecture but trains only on the permeability mismatch. Our analysis included (i) a base case with dense monitoring and moderate heterogeneity, (ii) eight data scenarios that combine variations in the size of the training set, the number of monitoring points, and correlation length, and (iii) rare event regimes constructed using an efficient tail sampling strategy.

Across all scenarios, the physics-informed model consistently produces permeability fields that yield more accurate pressure inferences while maintaining comparable accuracy in permeability inference.

In the base case, both models reproduce the true permeability field with comparable visual quality and similar $L_2$-errors, but the resulting pressure fields differ significantly. The data-driven model, despite its good permeability match, yields almost double pressure errors, compared to the physics-informed model. When we evaluate them on 1,000 test samples, a statistical analysis and visualizations confirm the higher accuracy in physics-informed inferences in most of the test cases. At the same time, the permeability errors remain similar between the two approaches.
These results demonstrate that the matching of permeability alone does not guarantee accurate pressure inferences. The physics-informed training with pressure loss in the learning yields flow responses that are more consistent with the underlying physics and the observed pressures.

The eight training scenarios provide a broader assessment of the robustness and generalization of the proposed method under changes in (i) training dataset size (5,000 vs.\ 50,000 samples), (ii) number of monitoring points (50 vs.\ 200), and (iii) correlation length of the permeability field (10~m vs.\ 100~m). These  scenarios are considered to understand the effects of data availability and heterogeneity on inverse-model performance. Across all scenarios, permeability inference errors for physics-informed and data-driven models are very close. This confirms that incorporating pressure loss into the training objective does not degrade the ability of the network to reproduce the  permeability, even when the training data become sparse or the system is highly heterogeneous. In contrast, pressure errors show much larger differences. For all eight scenarios, the physics-informed model outperforms the data-driven model, with the physics-informed model achieving lower pressure error in almost all of the realizations. The relative reduction in mean pressure error is significant compared to the data-driven method. 

The extreme data cases were studied to evaluate model performance in a regime that is particularly relevant for subsurface risk assessment, where high-consequence responses such as overpressure at critical locations, fault reactivation, or caprock failure lie in the tails of the system response distribution. To represent this regime, we generated extreme-data samples using an efficient tail-sampling approach following \citet{tong2021extreme}, which produced 6,200 samples exceeding a prescribed pressure threshold at a critical reservoir location. We then designed three extreme-data training and evaluation strategies and compared them with the base-case model trained on bulk samples. Models trained only on bulk data perform poorly on extreme-data samples, with substantially larger pressure inference errors than those observed in the bulk regime. This highlights a key limitation of bulk-trained surrogates: good average performance across the dominant data distribution does not ensure reliable accuracy in the tail region that governs risk. Adding extreme data samples to the training set reduces the error range for both the data-driven and physics-informed models, but the reduction is nearly twice as large for the physics-informed model as for the data-driven model. When both training and evaluation are restricted to extreme data samples, the physics-informed model attains pressure inference accuracy comparable to that of the base-case model in the bulk regime, despite being trained only on tail samples, whereas the data-driven model continues to exhibit larger errors. These findings suggest that the physics-informed framework is particularly effective for developing high-fidelity surrogates in extreme-data regimes, where accurate representation of tail behavior is critical for risk-sensitive decision-making.

\subsection{Computational considerations and practical implications}

Embedding a differentiable simulator in the training loop introduces additional computational cost relative to purely data-driven training, since each optimization step requires evaluating and differentiating through the forward model. In the present study, we use a two-dimensional steady-state single-phase flow model as a controlled setting that keeps the computational cost manageable while allowing the methodological advantages of simulator-embedded learning to be isolated clearly. However, the general framework is not restricted to this simplified setting. The same type of physics-informed learning can also be coupled to more complex differentiable simulators, including transient and multiphase flow models \cite{ur2026differentiable}.

Despite the additional training cost, the results suggest that improved physical fidelity and generalization can justify the use of simulator-embedded learning in applications where pressure predictions must remain reliable under uncertainty. In wastewater injection and CO$_2$ or H$_2$ storage projects, decisions such as allowable injection rates, monitoring-well placement, and assessment of leakage pathways depend critically on accurate pressure inference under uncertain geological conditions.

In practice, hybrid strategies may provide an effective balance between computational efficiency and physical accuracy. For example, a model may first be pretrained using the data-driven objective and then fine-tuned using the simulator-embedded physics-informed loss.

\subsection{Limitations and directions for future work}

While the proposed framework shows clear advantages over data-driven surrogates in the test problems considered here, several limitations remain.
\begin{itemize}
    \item The physics model employed in this study is a steady-state single-phase flow. This choice provides a controlled setting for isolating the effect of simulator-embedded learning, but many storage applications involve additional complexities such as multiphase flow and time dependence. Extending the present framework to such settings is an important direction for future work.

    \item While we demonstrated improved performance of the physics-informed model using synthetic data and the tail-sampling approach, field-scale deployment will require careful validation against real operational data.
    
\end{itemize}

\section{Conclusion}
In this work, we developed a physics-informed machine learning framework for reservoir characterization by embedding a differentiable full-physics flow simulator directly into the training loop of a neural network. This simulator-in-the-loop design allows the inverse model to be constrained not only by permeability mismatch, but also by the requirement that the inferred permeability field reproduce the observed pressure response. As a result, the learned surrogate is guided by the governing flow physics rather than by statistical fitting alone.

Across the numerical experiments considered here, this additional physical constraint consistently improved pressure inference accuracy relative to a data-driven baseline, while maintaining comparable accuracy in the recovered permeability field. The advantage was especially pronounced in settings with sparse observations, strong heterogeneity, and extreme-data regimes, where pressure responses are most difficult to predict and most important for risk assessment. These results indicate that incorporating a differentiable simulator can substantially improve the reliability and robustness of inverse models for subsurface flow applications.

The main trade-off of the proposed approach is its higher computational cost during training, since it requires repeated evaluation and differentiation through the forward simulator. Overall, the results suggest that simulator-embedded learning offers a promising path toward more physically faithful and trustworthy surrogate models for reservoir characterization. Such models are particularly relevant for applications such as geothermal energy extraction, CO$_2$, H$_2$ and waste water injection, where robust pressure inference under uncertainty is essential for operational decision-making, risk management, and long-term storage security.

\section*{Acknowledgments}
This material is based upon work supported by the U.S. Department of Energy, Office of Science Energy Earthshot Initiative as part of the project “Learning reduced models under extreme data conditions for design and rapid decision-making in complex systems" under Award number no. DE-SC0024721. DO was partially supported by U.S. Department of Energy, Office of Science (Basic Energy Sciences) Early Career Award number ECA1.
\section*{Conflict of interest disclosure}
The authors declare there are no conflicts of interest for this manuscript.
\section*{Data availability}
The datasets generated and/or used during the current study are publicly available in the following repository:

[\url{https://github.com/hrashid10/Extreme-Reservoir-Characterization}] 
\bibliography{citations} 

\end{document}